\documentclass[lettersize,journal]{IEEEtran}
\usepackage{times,latexsym}
\usepackage{amsmath,amsfonts}
\usepackage{algorithmic}
\usepackage{algorithm}
\usepackage{array}
\usepackage[caption=false,font=normalsize,labelfont=sf,textfont=sf]{subfig}
\usepackage{textcomp}
\usepackage{stfloats}
\usepackage{url}
\usepackage{verbatim}
\usepackage{graphicx}
\usepackage{cite}

\usepackage{amssymb}
\usepackage{enumerate}
\usepackage{multirow}
\usepackage{color}
\usepackage{booktabs}
\usepackage{multicol}
\usepackage{caption}

\usepackage[numbers]{natbib}
\usepackage{cleveref}
\crefname{section}{§}{§§}
\Crefname{section}{§}{§§}
\definecolor{newgreen}{RGB}{2,180,87} 

\begin{document}

\title{Unifying Structure Reasoning and Language Pre-training for Complex Reasoning Tasks}

\author{
Siyuan Wang, Zhongyu Wei, Jiarong Xu, Taishan Li, Zhihao Fan
\thanks{Manuscript received March 30, 2023. (Corresponding author: Zhongyu Wei).}
\thanks{Siyuan Wang, Zhongyu Wei and Zhihao Fan are with the School of Data Science, Fudan University, Shanghai 200433, China. And Zhongyu Wei is also with the Research Institute of Intelligent and Complex Systems, Fudan University, Shanghai 200433, China
(email: wangsy18@fudan.edu.cn; zywei@fudan.edu.cn; fanzh18@fudan.edu.cn).}
\thanks{Jiarong Xu is with the School of Management, Fudan University, Shanghai 200433, China (jiarongxu@fudan.edu.cn).}
\thanks{Taishan Li is with the School of Information Science and Technology, ShanghaiTech University, Shanghai, 201210, China (litsh@shanghaitech.edu.cn).}
}

\markboth{IEEE/ACM TRANSACTIONS ON AUDIO, SPEECH, AND LANGUAGE PROCESSING, 2023}%
{\MakeLowercase{\textit{et al.}}: Unifying Structure Reasoning and Language Model Pre-training for Complex Reasoning}


\maketitle

\begin{abstract}
Recent pre-trained language models (PLMs) equipped with foundation reasoning skills have shown remarkable performance on downstream complex tasks. However, the significant structure reasoning skill has been rarely studied, which involves modeling implicit structure information within the text and performing explicit logical reasoning over them to deduce the conclusion. This paper proposes a unified learning framework that combines explicit structure reasoning and language pre-training to endow PLMs with the structure reasoning skill. It first identifies several elementary structures within contexts to construct structured queries and performs step-by-step reasoning along the queries to identify the answer entity. The fusion of textual semantics and structure reasoning is achieved by using contextual representations learned by PLMs to initialize the representation space of structures, and performing stepwise reasoning on this semantic representation space. Experimental results on four datasets demonstrate that the proposed model achieves significant improvements in complex reasoning tasks involving diverse structures, and shows transferability to downstream tasks with limited training data and effectiveness for complex reasoning of KGs modality.
\end{abstract}


\section{Introduction}
\IEEEPARstart{R}{ecent} years have witnessed an ever-growing research on complex reasoning, which requires comprehending the given information and applying complex rules to draw inferences~\cite{songer2009and, hassabis2017neuroscience, wang2022lsat, czechowski2021subgoal}. As a defining property of advanced intelligence, it inspires immense potential for numerous real-world applications, such as fact checking~\cite{thorne2018fever, ostrowski2020multi}, math word problem solving~\cite{dua2019drop, amini2019mathqa}, natural language navigation~\cite{chen2019touchdown, kim2020arramon} and medical diagnosis~\cite{datta2020hybrid, van2021clinical}.
Existing large-scale pre-trained language models (PLMs) have shown superior performance on various downstream tasks, exhibiting the general linguistic reasoning skill for understanding contextual information learned from broad data~\cite{devlin2018bert, liu2019roberta, lan2019albert, he2020deberta}. However, complex reasoning tasks are far more challenging and diverse, involving further foundation reasoning skills. 
\begin{figure}[!th]
\centering
\includegraphics[width=1.0\columnwidth]{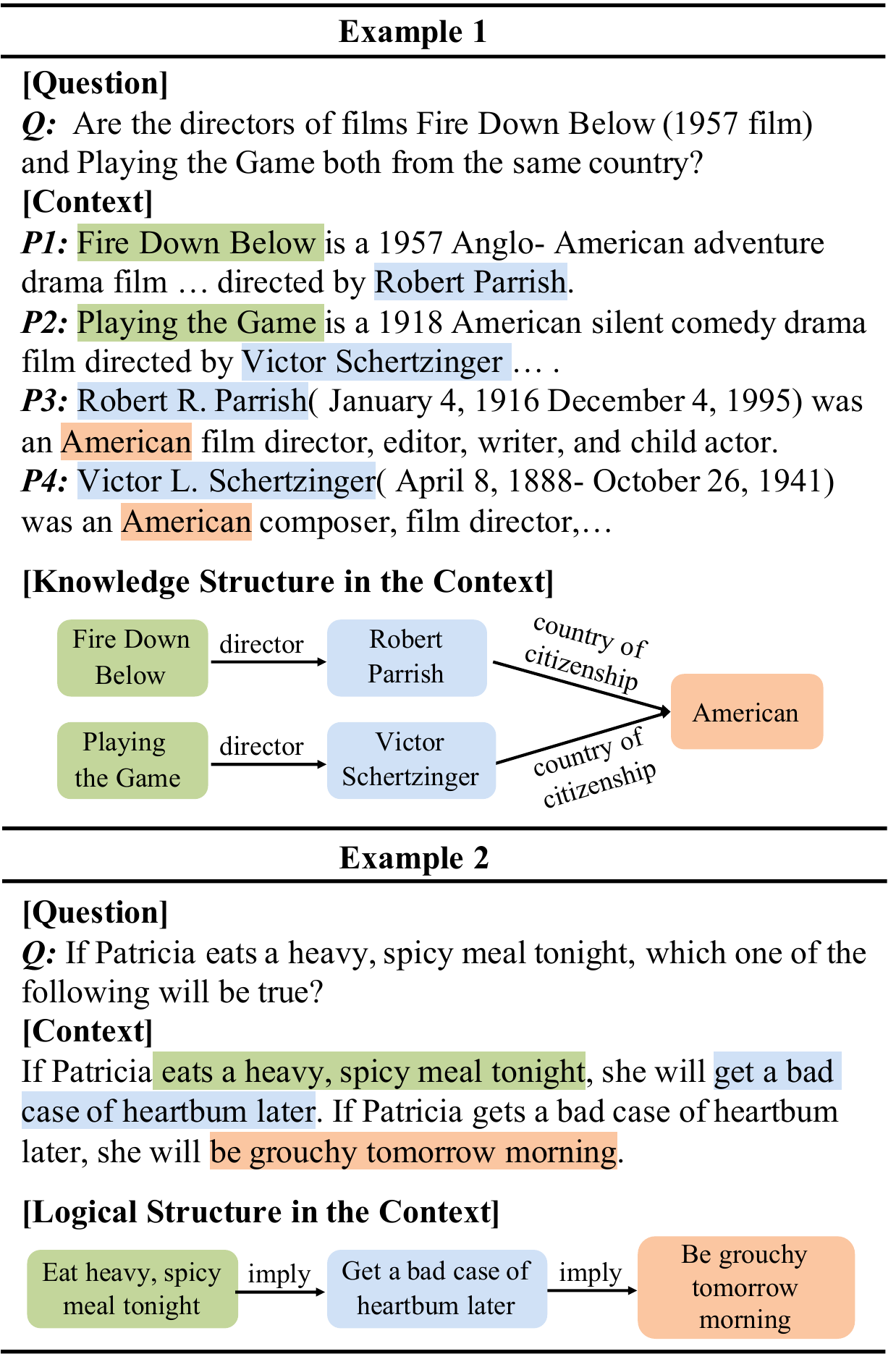}
\caption{\label{figure1} A multi-hop reasoning example (Example 1) from~\cite{ho-etal-2020-constructing} and a logical reasoning example (Example 2) from~\cite{yu2020reclor}.
The contexts embody implicit structures for reasoning. 
The shaded rectangles are entities and the texts on top of arrows are relations in the knowledge and logical structures, and the shaded phrases in contexts are the mentions of corresponding entities. 
}
\end{figure}
For example, the numerical reasoning skill to abstract quantitative information from text~\cite{geva2020injecting, petrak2022improving}, and the spatial reasoning skill to perceive spatial relations between objects are two essential abilities~\cite{moen2020strengthening, mirzaee2021spartqa, subramanian2022reclip}.
Equipping PLMs with these foundation skills enables them to be adapted to a wide variety of downstream reasoning tasks, which significantly steps forward in artificial general intelligence.

In addition to these foundation reasoning skills, there is also an essential ability that is rarely explored, namely the structure reasoning skill.
Structure reasoning aims to model implicit structure information within the text and perform explicit logical reasoning over them to deduce the conclusion. 
In Figure~\ref{figure1}, we show two complex reasoning examples~\cite{yang2018hotpotqa,welbl2018constructing,yu2020reclor,liu2020logiqa} that involve the structure reasoning skill. We can see that contexts embody rich information for answering these questions. 
Although the relevant information typically exists in complex and sparse forms (shaded phrases) in context, they are implicitly organized as a structure. Specifically, the contexts of Example 1 (a multi-hop reasoning example) and Example 2 (a logical reasoning example) respectively embody an intrinsic knowledge structure and logical structure. 
Through explicit reasoning over these structures, the correct answers can be inferred. We argue that formulating relevant information from unstructured text as topological structures and performing logical reasoning over them can help solve extensive complex questions. In this paper, we thus explore endowing the PLMs with the foundation structure reasoning skill for complex reasoning.

Previous research usually generates large-scale question-answer pairs that require reasoning and continue training PLMs over the synthetic data to inject foundation reasoning skills~\cite{li2019teaching, yoran2021turning, zhang2022skillnet}.
Such data-driven methods attempt to learn foundation skills in an implicit manner, making it uncertain whether they really gain these capabilities or just exploit data biases as shortcuts for question answering.
Instead, we aim to model the structure reasoning skill more explicitly and take inspiration from geometric embedding methods for step-by-step reasoning over structures (knowledge graphs)~\cite{hamilton2018embedding, ren2020query2box, ren2020beta}. 

To this end, we propose a unified learning framework to combine explicit structure reasoning and language pre-training. For structure reasoning, we first define several elementary structures within contexts and perform reasoning over the structure for the answer entity identification. Specifically, our model first utilizes these context-inherent basic structures to construct structured queries with corresponding answer entities, and encode them as geometric shapes in a representation space.
Then the reasoning along the query structure is explicitly conducted by iteratively executing logical operations over the representation space. 
The goal is to push the answer representation to be close enough to the final query representation.
For the fusion of textual semantics and structure reasoning, we use contextual representations learned by PLMs from contexts to initialize the representation space of structures, and perform stepwise reasoning on this semantic representation space.
In this way, the pre-trained model is taught with the explicit structure reasoning capability over text and can easily generalize to different structures composed of these studied elementary ones.

We conduct experiments on two datasets for multi-hop reasoning, HotpotQA~\cite{yang2018hotpotqa} and 2WikiMultiHopQA~\cite{ho-etal-2020-constructing}, as well as two logical reasoning datasets, ReClor~\cite{yu2020reclor} and LogiQA~\cite{liu2020logiqa}.
The results demonstrate that our model unifying the explicit structure reasoning skill into PLMs achieves significant improvements in complex reasoning tasks involving diverse structures. Further analysis shows its transfer ability to downstream tasks with limited training data and effectiveness for complex reasoning of KGs modality. The contributions of this work can be summarized as follows:
\begin{itemize}
    \item We propose a new foundation reasoning skill, namely structure reasoning skill, to formulate implicit
    structure information from the text and perform logical reasoning over them, which is significant for complex reasoning.
    \item We propose a unified learning framework to explicitly inject the structure reasoning skill into PLMs for better generalizing to different complex reasoning tasks with only unstructured text but involving structure reasoning. It is a generic framework that can be plugged into different pre-trained language models.
    \item We present extensive experiments, demonstrating the effectiveness and low-resource transfer ability of our proposed framework. 
\end{itemize}

\section{Related Work}
\subsection{Foundation Reasoning Skills Learning}
Large-scale pre-trained language models~\cite{devlin-etal-2019-bert, liu2019roberta, lan2019albert, brown2020language} facilitate a variety of downstream NLP applications but have difficulty in challenging and diverse complex reasoning tasks. To generally improve complex reasoning, a set of foundation reasoning skills is introduced for capturing some common reasoning abilities across different tasks and are incorporated into PLMs. For example, \cite{geva2020injecting, petrak2022improving} propose to inject numerical reasoning skills for complex tasks requiring understanding and reasoning over quantitative information. Commonsense reasoning~\cite{lin2019commongen} and logical inference abilities~\cite{pi2022logigan} are also taught to pre-trained language models for solving problems involving commonsense and informal logic. Here, we focus on learning another significant structure reasoning skill, which aims to formulate the implicit structure from text and perform explicit reasoning over them. 
Even recent large language models, such as GPT-3~\cite{brown2020language} and ChatGPT, are yet to master explicit structural reasoning. These models continue to have difficulties in complex reasoning tasks, such as multi-hop reasoning, and are prone to generating factual errors. This highlights the crucial need to explore and develop structural reasoning skills.

Existing work primarily exploits pre-defined templates to synthesize substantial amounts of data requiring different reasoning skills, which are in the form of question-answer pairs or mask-out statements. They endow PLMs with these foundation reasoning skills by further refining PLMs on augmented training data. Different from this implicit teaching process, we inject the structure reasoning skills into PLMs more explicitly by stepwise performing logical operations along structures over semantic representation space.

\subsection{Knowledge Enhanced Pre-trained Language Models}
Another line of related work is on incorporating structured knowledge from external sources into PLMs to provide extra clues for reasoning~\cite{talmor2020olmpics, lin2019commongen, yang2019enhancing, wang2020k}.
The main challenge is that the infused knowledge and the text usually have heterogeneous representation space. To integrate heterogeneous information, some works need to revise the architecture of PLMs with an additional fusion module~\cite{zhang-etal-2019-ernie, yu2020jaket}. Others require hand-crafted heuristics to combine knowledge and text into a unified data structure, such as a sequence~\cite{guan2020knowledge}, a sentence tree~\cite{liu2020k} and a word-knowledge graph\cite{sun2020colake}, and convert them into input sequences for encoding. However, these methods ignore the structural information of knowledge graphs in some cases~\cite{lu2021kelm}. Besides, knowledge retrieval is required to identify relevant information from external sources and the inaccurate retrieval procedure introduces noisy knowledge inevitably. 

Contrary to these works incorporating external structured information, we study modeling the intrinsic structures in the text and reasoning over them. It is essential for complex reasoning tasks~\cite{yang2018hotpotqa, welbl2018constructing, yu2020reclor} where the question needs to be answered grounded on the context with complex but sparse information, which actually constitutes an implicit structure. Although some attempts can be viewed as structure modeling in text, they only consider reasoning over simple triplets~\cite{sun2019ernie, joshi2020spanbert,qin2020erica, sun2021ernie}.
Moreover, they mainly adopt span-based masked pre-training or transform the triplets to text sequences for contrastive learning, which ignores the structured information and explicit structure reasoning over them. The crucial difference is that we propose to unify language modeling and explicit structure reasoning on both simple and complex structures within text. 

\begin{figure*}[!th]
\centering
\includegraphics[width=2.0\columnwidth]{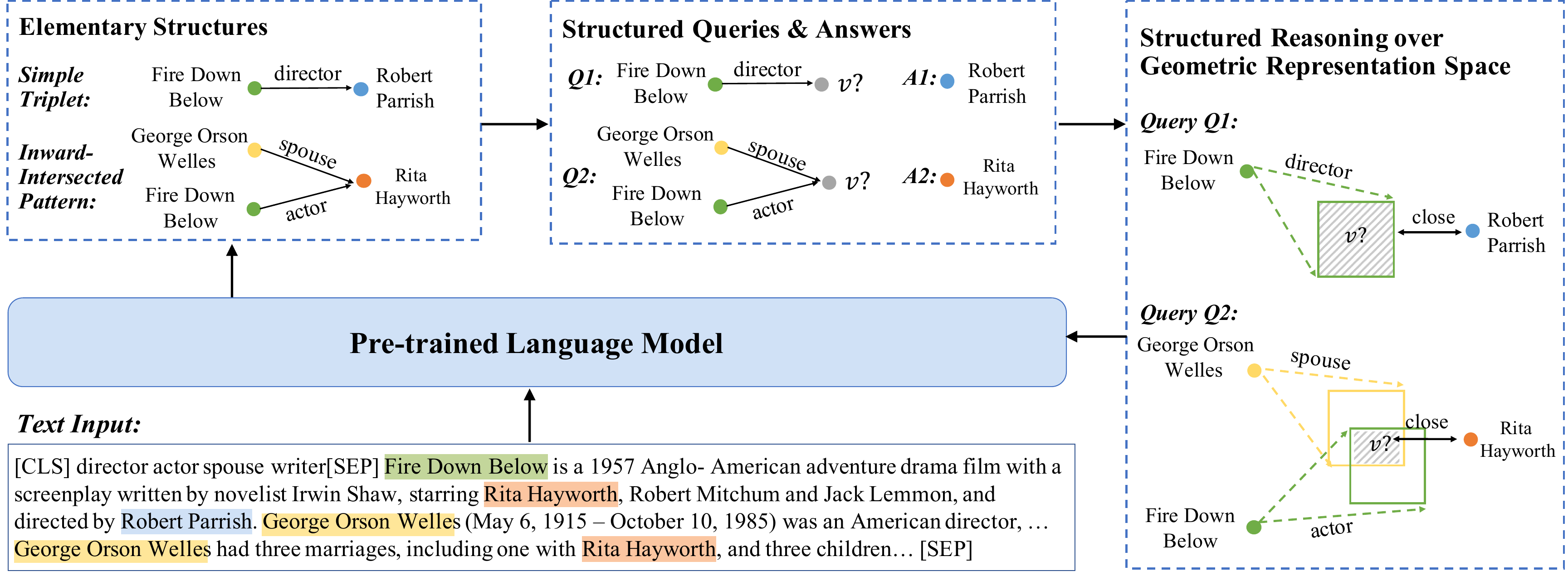}
\caption{\label{figure_framework} The overall architecture of our framework unifying structure reasoning and language modeling.}
\end{figure*}
\section{Methodology}
In this work, we propose to unify explicit structure reasoning and language pre-training in one framework for complex reasoning (see Figure~\ref{figure_framework}). 
Given a piece of text sequence $s$ with existing entities $\mathcal{E}=\{e_1, e_2, e_3, ...\}$ and their relations $\mathcal{R}=\{...,r_{i,j},...\}$ constituting triplets $\mathcal{T}=\{(e_i, r_{i,j}, e_j)\in\mathcal{E}\times\mathcal{R}\times\mathcal{E} \}$, our goal is to learn the contextual representations of $s$ together with reasoning over complex structure information within the text.
Complex structures vary drastically in different texts, which usually exhibit in the form of multi-step paths and intersected triplets, or a combination of them. Therefore, we aim to learn general reasoning skills over elementary structures to achieve generalizability to any structures.
We define four types of elementary structures that can constitute almost arbitrary complex structures by free combination. It includes the simple triplet and three complex structures, i.e., the two-step path, outward-intersected pattern, and inward-intersected pattern.

In this section, we first introduce how to obtain the representations of entities and relations from text, which are the basic units of our defined structures (\cref{sec:knowledge_representation}). We then detailedly describe these elementary structures and their identification process from the text (\cref{sec:knowledge_structures}). We follow the geometric embedding-based method for explicit structure reasoning over the semantic representation space(\cref{sec:structured_reasoning}). Finally, we show the pre-training process to learn a structure-aware language representation for acquiring structure reasoning skills and its fine-tuning stage on downstream complex reasoning tasks (\cref{sec:pretraining_finetuning}).

\subsection{Basic Structure Unit Representations}
\label{sec:knowledge_representation}
Taking the text sequence $s$ with tokens $\{s_1, s_2, ..., s_{|s|}\}$ as input, the pre-trained language model outputs a sequence of contextual representations $H=\{h_1, h_2, ..., h_{|s|}\}$. We do not maintain separate embeddings for structures. Instead, we obtain representations of basic structural units from the contextual language representations, to fuse textual and structured semantics and improve the structure reasoning skill during language pre-training. In detail, for an entity $e_i$ in the text $s$, we take the average hidden state of its start token $s_{start}^{e_i}$ and end token $s_{end}^{e_i}$ as its representation $\bf e_i$. If the entity occurs in the text sequence multiple times $N_{e_i}$, we further take the average of its multiple occurrences as Eq.~\ref{entity_rep}.
As the relation may not be explicitly or consecutively mentioned in the text, we additionally concatenate the relations of all existing triplets with the text sequence $s$ as the input \texttt{[CLS] $r_{i,j}$, ... [SEP] $s$ [SEP]} for encoding. For each relation $r_{i,j}$ we also get the average hidden state of its start token $s_{start}^{r_{i,j}}$ and end token $s_{end}^{r_{i,j}}$ as its representation $\bf r_{i,j}$. 
\begin{align}
    &{\bf e_i}=\frac{1}{N_{e_i}}\sum_1^{N_{e_i}} \frac{1}{2}(h^{e_i}_{start}+h^{e_i}_{end}) \label{entity_rep}\\
    &{\bf r_{i,j}} = \frac{1}{2} (h^{r_{i,j}}_{start}+h^{r_{i,j}}_{end}) \label{relation_rep}
\end{align}

\subsection{Elementary Structures Identification}
\label{sec:knowledge_structures}
\begin{figure}[!th]
\centering
\includegraphics[width=0.92\columnwidth]{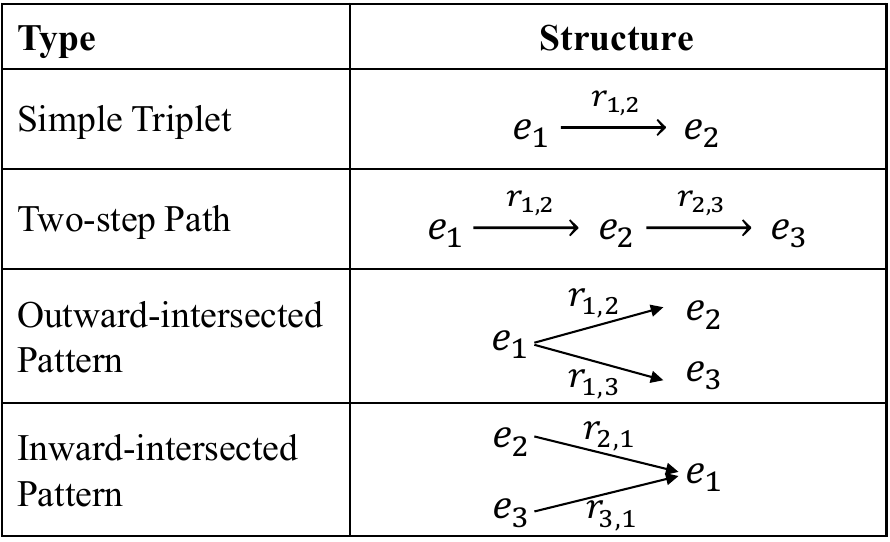}
\caption{\label{figure_knowledge} The definition of different elementary structures.
}
\end{figure}

After getting the basic unit representations, we propose four types of elementary structures for reasoning over both simple and complex structures in the text as shown in Figure~\ref{figure_knowledge}. A simple triplet is comprised of two entities and one relation. Complex structures can be multi-step paths, intersected triplets, or a combination of them. We consider three elementary complex structures, including the two-step path, outward-intersected pattern and inward-intersected pattern.
We aim to learn general reasoning skills over these four types of elementary structures to achieve generalizability to model any complex structures in text, even those never seen during pre-training. 
For example, the complex knowledge structure of the first example in Figure~\ref{figure1} 
can be modeled by combining several two-step paths and an inward-intersected pattern.

Taking simple triplets extracted from the context as input, we further identify the other three complex structures from the text. For two triplets $(e_1, r_{1,2}, e_2)$ and $(e_2, r_{2,3}, e_3)$ appear in the same context, if the tail $e_2$ of the first triplet is the head of the other, we treat it as a \emph{two-step path}. If two triplets $(e_1, r_{1,2}, e_2)$ and $(e_1, r_{1,3}, e_3)$ in the same context share the same head $e_1$ but their tails are different, we combine them as a \emph{outward-intersected 
pattern}. If two triplets $(e_2, r_{2,1}, e_1)$ and $(e_3, r_{3,1}, e_1)$ have the same tail $e_1$ but differ in heads, we recognize them as a \emph{inward-intersected pattern}. 

\subsection{Structure Reasoning Injection}
\label{sec:structured_reasoning}
Based on these identified structures, we can construct corresponding structural queries and perform explicit structure reasoning along them to reach answers. Inspired by~\cite{ren2020query2box, ren2020beta}, we utilize geometric embedding learning methods for stepwise updating structural query representation and encourage the query representation to be close to the answer representation in the vector space. We use contextual language representations of structures from PLMs to initialize their geometric embeddings to integrate textual and structured semantics. 
\begin{figure}[!th]
\centering
\includegraphics[width=0.92\columnwidth]{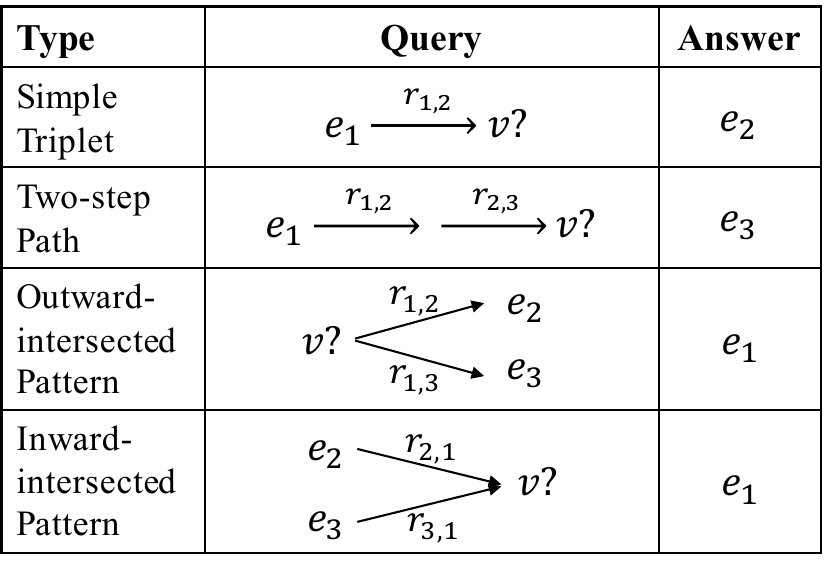}
\caption{\label{figure_queries} Structured query construction for different structures and $v?$ is the query node.
}
\end{figure}

\paragraph{Structural Query Construction} We construct structural queries on top of pre-defined elementary structures by taking different entities from different structures as target answers (query node) and viewing the remaining as queries. Specifically, 
the tail entity is extracted as the target in the simple triplet, while the intersected entities are treated as answers for intersected patterns. For the two-step path, we remove the intermediate entity and take the last entities as the query target. The detailed strategies for structured query construction with their answer entities are shown in Figure~\ref{figure_queries}.

\paragraph{Geometric Representation Learning} As a structure reasoning method for answering complex queries over knowledge graphs, it models a set of entities into a geometric region in the vector space. 
Correspondingly, it learns the logical operators in queries including relation projection and intersection as operations over geometric regions, which will result in new regions.
The structure reasoning can be conducted from the start entities by explicitly executing the logical operators along the structural query to update the query representation, and encourage the answer entities to be inside or close to the final region of the query.

Specifically, we use boxes (hyper-rectangles) as the geometric regions to represent the structural queries.
A box embedding $\bf b$ is composed of a center vector $\text{Cen({\bf b})}$ and an offset vector $\text{Off({\bf b})}$ as $\bf b = (\text{Cen({\bf b})}, \text{Off({\bf b})})$, and model a set of entities in the following box:
\begin{align}
    \{{\bf e}: \text{Cen({\bf b})}-\text{Off({\bf b})} \preceq {\bf e} \preceq \text{Cen({\bf b})}+\text{Off({\bf b})} \}
\end{align}
where $\text{Cen({\bf b})}$ and $\text{Off({\bf b})}$ are respectively the center and positive offset of the box. Each entity $e$ can then be represented as a zero-offset box ${\bf e}=(\text{Cen({\bf e})}, {\bf 0})$ with only one element as the center. The relation $r$ is also embedded as $\bf r = (\text{Cen({\bf r})}, \text{Off({\bf r})})$ and the relation projection operator is modeled as ${\bf b}+{\bf r}$ to respectively sum the centers and offsets, and obtain a transformed box embedding. The intersection operator over multiple box embeddings $\{ {\bf b_1}, {\bf b_2}, ... {\bf b_n}\}$ is modeled as $\bf b_{\cap} = (\text{Cen}(b_{\cap}), \text{Off}(b_{\cap}))$, where $\text{Cen}(\bf b_{\cap})$ and $\text{Off}(\bf b_{\cap})$ are computed as Eq.~\ref{inter_center} and~\ref{inter_offset}. $a_i$ is the attention weight over the box center $\bf b_i$ and $\text{DeepSets}(\cdot)$ is the permutation-invariant function over sets 
~\cite{zaheer2017deep}.
\begin{align}
    \text{Cen}({\bf b_{\cap}}) &= \sum^n_{i} a_i \odot \text{Cen}({\bf b_i}) \label{inter_center} \\
    \text{Off}({\bf b_{\cap}}) &= \text{Min}(\{\text{Off}({\bf b_i}), i\in 1,...,n\}) \nonumber \\
    &\odot \sigma(\text{DeepSets}(\{{\bf b_1}, i\in 1,...,n\})) \label{inter_offset}
\end{align}

\paragraph{Structure Reasoning} 
According to the constructed structural queries, we then perform structure reasoning using the geometric representation learning method and make the answer embeddings to be as similar as possible to the query box embedding. We take the contextual semantic representations of entities and relations in Eq.~\ref{entity_rep} and~\ref{relation_rep} as their center vectors $\text{Cen({\bf e})}$ and $\text{Cen({\bf r})}$. We do not learn an adaptive offset for different relations as in~\cite{ren2020query2box} but train a shared one to reduce the extra learning burden and fuse the structure reasoning into language pre-training more 
smoothly. The loss of structure reasoning for a query-answer pair using negative sampling is calculated as follows:
\begin{align}
    \mathcal{L}_{QA}=&-\mathop{\log}\sigma(\gamma-d({\bf a}, {\bf q})) \nonumber \\
    &- \sum_{k=1}^K \frac{1}{K}\mathop{\log}\sigma(d({\bf a^{'}_k}, {\bf q})-\gamma)
\end{align}
where $\bf q$ is the derived box embedding of the query $q$ and $\bf a$ is the answer entity embedding of $a$. $\bf a^{'}_k$ is the embedding of a negative answer span $a^{'}_k$ randomly sampled from the same text. $\gamma$ is the fixed margin and the function $d(\cdot, \cdot)$ measures the distance between an entity and a box. 

The detailed distance calculation between an entity embedding $\bf e$ and a query box embedding $\bf b = (\text{Cen({\bf b})}, \text{Off({\bf b})})$ consist of two parts~\cite{ren2020query2box} as following:
\begin{align}
    d({\bf e}, {\bf b}) = d_{out}({\bf e}, {\bf b}) + \alpha \ d_{in}({\bf e}, {\bf b}) 
\end{align}
where $\ d_{out}({\bf e}, {\bf b})$ and $d_{in}({\bf e}, {\bf b})$ are respectively the outer distance between the entity and the closest box corner and the inner distance between the box center and one box corner. $\alpha$ is a scalar coefficient balancing these two distances. 
\begin{align}
    d_{out}({\bf e}, {\bf b}) = &||\max({\bf e}-{\bf b}_{\max}, {\bf 0}) + \nonumber \\
    & \max({\bf b}_{\min}-{\bf e}, {\bf 0})||  \\
    d_{in}({\bf e}, {\bf b}) = &||\text{Cen({\bf b})} - \nonumber \\
    & \min({\bf b}_{\max}-\max({\bf b}_{\min}, {\bf e}))|| \\
    {\bf b}_{\max} = &\text{Cen({\bf b})}+\text{Off({\bf b})} \\
    {\bf b}_{\min} = &\text{Cen({\bf b})}-\text{Off({\bf b})}
\end{align}

\subsection{Pre-training \& Fine-tuning}
\label{sec:pretraining_finetuning}
In order to inject the structure reasoning skill into PLMs, we jointly optimize structure reasoning and language modeling. 
We follow the masking strategy of RoBERTa~\cite{liu2019roberta} for masked language modeling.
The overall pre-training objective is the combination of the structure reasoning loss $\mathcal{L}_{SR}$ and the masked language modeling loss $\mathcal{L}_{MLM}$ as Eq.~\ref{loss}.
\begin{align}
    \mathcal{L} = \mathcal{L}_{SR} + \mathcal{L}_{MLM} \label{loss}
\end{align}
For each text, we do not model all existing elementary structures. We have attempted to model different numbers of complex structures which all show similar performance on downstream tasks. However, modeling more complex structures simultaneously will increase the computational burden and slow down the optimization speed.
Therefore, we randomly choose a simple triplet and one of the other three complex structures to construct both a simple query and a complex query for structure reasoning over text. The $\mathcal{L}_{SR}$ is computed as follows:
\begin{align}
    \mathcal{L}_{SR} = \lambda_1 \mathcal{L}^{\text{simple}}_{QA} + \lambda_2\mathcal{L}^{\text{complex}}_{QA}
\end{align}
where $\mathcal{L}^{\text{simple}}_{QA}$ and $\mathcal{L}^{\text{complex}}_{QA}$ are respectively the structure reasoning loss over simple and complex structures queries. $\lambda_1$ and $\lambda_2$ are weighted hyper-parameters to balance the losses.

With the structure reasoning skill injected during language pre-training, our model learns structure-aware language representations and can be directly fine-tuned on downstream language tasks requiring complex structure reasoning.

\section{Experiments}
In the experiments, we evaluate our model on the textual complex reasoning tasks, and further analyze the contribution of each structural component and the transfer ability to downstream tasks with limited training data. Moreover, the capabilities of our model in complex reasoning over KGs and knowledge probing are also validated.


\subsection{Experimental Setup}
\label{implemental}
\paragraph{Pre-training phase} We utilize Wikipedia documents with entities and their relations constructed by~\cite{qin2020erica} to obtain structures for structure reasoning enhanced language pre-training. Specifically, it utilizes spaCy to perform named entity recognition. It then aligns entity mentions to Wikidata items for annotating relations. According to their human evaluation, the F1 scores of entity and relation extraction are respectively 84.7\% and 25.4\%.

We concatenate all documents and splits them into sequences of the same length for masked language modeling so that each sequence may cover multiple documents. Then we can utilize triplets from different documents involved in one sequence to construct cross-document structures and queries involving complex reasoning.
We take RoBERTa-base~\cite{liu2019roberta} as the backbone of our language model. We set the learning rate as 5e-5. The hyper-parameters $\lambda_1$ and $\lambda_2$ that balance structure reasoning losses over different structures are set as $1$ and $0.1$, respectively. The fixed margin $\gamma$ is set as 24. The coefficient to balance outer and inner distances between an entity and a query box is set as $\alpha=0.02$.
\begin{table*}[ht]
\setlength\tabcolsep{12pt}
\begin{center}
\caption{Experimental results of different pre-trained models on HotpotQA and 2WikiMultiHopQA datasets. Results marked with * indicate the percentage of sampled questions that can be correctly answered by ChatGPT.  
}
\label{table_result_multihop}
\resizebox{0.73\textwidth}{!}{
\begin{tabular}{lcccccc}
\toprule
\multirow{2}{*}[-0.7ex]{\centering Model}
&\multicolumn{2}{c}{Answer} &\multicolumn{2}{c}{Supporting Fact} &\multicolumn{2}{c}{Joint} \\
\cmidrule(lr){2-3}\cmidrule(lr){4-5}\cmidrule(lr){6-7}
 & EM & F1 & EM & F1 & EM & F1 \\
\midrule
\multicolumn{7}{c}{ \ \ \ \ HotpotQA} \\
\midrule
\emph{RoBERTa}~\cite{liu2019roberta} & 64.82 & 78.68 & 61.18 & 86.61 & 42.66 & 70.12 \\
\emph{CoLAKE}~\cite{sun2020colake} & 63.31 & 77.93 & 61.56 & 86.73 & 42.41 & 69.63 \\
\emph{KEPLER}~\cite{wang2021kepler} & 63.20 & 77.29 & 61.74 & 86.82 & 42.41 & 69.07 \\
\emph{ERICA}~\cite{qin2020erica} & 64.34 & 78.52 & 61.50 & 86.64 & 43.63 & 70.13 \\
\emph{Ours(PTransE)} & \bf 65.25 & 79.07 & \bf 62.23 & 87.05 & 43.41 & 70.72 \\
\emph{Ours} & 64.85 & \bf 79.14 & 62.07 & \bf 87.06 & \bf 43.82 & \bf 70.87 \\
\midrule
\emph{ChatGPT} & 69.00* & - & - & - & - & - \\
\midrule
\multicolumn{7}{c}{ \ \ \ \ 2WikiMultiHopQA} \\
\midrule
\emph{RoBERTa}~\cite{liu2019roberta} & 62.29 & 66.25 & 80.30 & 90.68 & 55.20 & 62.56 \\
\emph{CoLAKE}~\cite{sun2020colake} & 62.50 & 66.80 & 80.01 & 90.53 & 55.25 & 62.99 \\
\emph{KEPLER}~\cite{wang2021kepler} & 62.25 & 66.83 & 79.87 & 90.50 & 54.83 & 63.00 \\
\emph{ERICA}~\cite{qin2020erica} & 66.16 & 71.22 & 79.82 & 90.66 & 57.76 & 67.14 \\
\emph{Ours(PTransE)} & 66.28 & 71.14 & 80.30 & 90.52 & 58.57 & 67.13\\
\emph{Ours} & \bf 66.30 & 71.11 & \bf 80.41 & \bf 90.70 & \bf 59.00 & \bf 67.32 \\
\midrule
\emph{ChatGPT} & 50.50* & - & - & - & - & - \\
\bottomrule
\end{tabular}
}
\end{center}
\end{table*}

\paragraph{Fine-tuning Period} We evaluate our model on different tasks requiring reasoning over text with implicit complex structures, 
including multi-hop reasoning on HotpotQA~\cite{yang2018hotpotqa} and 2WikiMultiHopQA~\cite{ho-etal-2020-constructing} datasets, and logical reasoning on ReClor~\cite{yu2020reclor} and LogiQA~\cite{liu2020logiqa} datasets.
For datasets without publicly available test sets, we randomly split the development set into two half-sections for validation and testing respectively.
We evaluate our model on the validation set of each dataset to choose parameters for testing. 
All models are implemented using Huggingface~\cite{wolf2019huggingface}.
For both HotpotQA and 2WikiMultiHopQA, the batch size is set to 96 while for ReClor and LogiQA, we use a batch size of 24. 
The Adam with $\beta_1 = 0.9$ and $\beta_2 = 0.98$ is taken as the optimizer and we use a linear learning rate scheduler with $10\%$ warmup proportion. 
The proposed systems and other comparison models are trained on NVIDIA Tesla V100 GPUs.


\subsection{Overall Performance}
\paragraph{Comparison Models} We compare our framework with the baseline model and several other structure-injected models, which can be divided into four categories: (1) \emph{RoBERTa} is the baseline model without any structure modeling. (2) \emph{CoLAKE}~\cite{sun2020colake} and \emph{KEPLER}~\cite{wang2021kepler}: external knowledge structure enhanced language models. \emph{CoLAKE} aggregates surrounding triplets and converts them into input sequences for entity masked pre-training. \emph{KEPLER} incorporates unnecessarily relevant triplets and learns a unified structure-textual representation. (3) \emph{ERICA}~\cite{qin2020erica} models the structure information in text including entity and relations through contrastive textual discrimination to improve language modeling. (4) \emph{Ours} is our proposed model which learns reasoning skills over intrinsic complex structures. 
We also compare a variant \emph{Ours(PTransE)} utilizing the path-based TransE method~\cite{lin2015modeling} for structure reasoning which embeds the query as single point instead of the box in vector space and only models simple triplets and multiple-step paths. 
Additionally, we compared our framework to the recent ChatGPT model by randomly sampling 200 instances from the test sets of each dataset.

\paragraph{Multi-hop Reasoning}
Multi-hop reasoning aims to aggregate multiple pieces of documents and formulate cross-document structures to answer a complex question. 
We adopt two textual multi-hop reasoning datasets, HotpotQA and 2WikiMultiHopQA, that involve various reasoning steps (2$\sim$4) and structures. HotpotQA and 2WikiMultiHopQA respectively consist of 90,447 / 7,405 / 7,405 and 167,454 / 3,702 / 3,703 samples in training, development and test sets.
They both need to identify an answer span to the question from the context and predict the supporting facts to explain the reasoning.
We concatenate the question and context and take them as input for fine-tuning. The models are trained on HotpotQA for 10 epochs with the learning rate 7e-5, and on 2WikiMultiHopQA for 5 epochs with the learning rate 3e-5. 

The experimental results are shown in Table~\ref{table_result_multihop}. 
We have the following findings. 
\begin{enumerate}[-]
    \setlength{\itemsep}{1pt}
    \setlength{\parskip}{1pt}
    \item Our model outperforms almost all other models on both HotpotQA and 2WikiMultiHopQA datasets.  Even when compared to the powerful ChatGPT, our model achieves comparable performance (Noting that ChatGPT can not predict the answer phrase exactly).
    This demonstrates the necessity of explicit structure reasoning over text during language modeling. 
    \item Compared to \emph{CoLAKE} and \emph{KEPLER} utilizing external knowledge structures to enhance language modeling, our models and \emph{ERICA} achieve a considerable improvement.
    This suggests that instead of introducing external knowledge, it is more important to directly model the intrinsic structures in text for complex multi-hop reasoning.
    \item To illustrate the utility of box embedding method, we compare \emph{Ours}  with \emph{Ours(PTransE)}. The results show that \emph{Ours} performs better than \emph{Ours(PTransE)} in the joint performance of answer prediction and supporting fact prediction, which shows that the box embedding method is more effective than path-based TransE for explicit complex structure reasoning. 

\end{enumerate}

\paragraph{Logical Reasoning}
Logical reasoning aims to uncover the logical structures within the text and perform inference over them to deduce the answer. We evaluate two benchmarks covering diverse logical structures, ReClor and LogiQA, which are collected from standardized exams including GMAT and LSAT and National Civil Servants Examination of China respectively. Reclor and LogiQA respectively contain 4,638 / 250 / 250 and 7,376 / 651 / 651 data points for training, validation and testing.
More specifically, we conduct multiple-choice question answering on ReClor and LogiQA by taking a context, a question and four options as the input and outputting the most plausible option as the answer.
We concatenate the context, the question and each option as an input sequence for encoding, resulting in four formulated sequences, and then choose the one with the highest score.
The maximum sequence length is set as 256, the number of training epochs is 10 and the learning rate is 1e-5. 
\begin{table}[!h]
\setlength\tabcolsep{8pt}
\begin{center}
    \caption{Evaluation accuracies (\%) of different pre-trained models on both validation and test sets of ReClor and LogiQA.}
    \label{table_result_logical}
    \resizebox{0.46\textwidth}{!}{
    \begin{tabular}{lcccc}
    \toprule
    \multirow{2}{*}[-0.7ex]{Model}
    &\multicolumn{2}{c}{ReClor} 
    &\multicolumn{2}{c}{LogiQA} \\
    \cmidrule(lr){2-3}\cmidrule(lr){4-5} & Val. & Test
    & Val. & Test \\
	\midrule  				
	\emph{RoBERTa} & 52.8 & 52.8 & 34.56 & 32.26 \\
	\emph{CoLAKE} & 51.6 & 50.4 & 34.72 & 31.18 \\  
	\emph{KEPLER} & 46.8 & 44.0 & 31.18 & 27.96 \\
	\emph{ERICA} & 55.6 & 55.2 & 31.18 & 26.27 \\
	\emph{Ours(PTransE)} & 52.8 & 56.4 & 31.34 & 32.87 \\
    \emph{Ours} & \bf 56.8 & \bf 57.6 & \bf 35.18 & \bf 33.03 \\
    \midrule
    \emph{ChatGPT} & - & 56.0 & - & 34.00 \\
	\bottomrule
 	\end{tabular}
 	}
	\end{center}
\end{table}

The results of logical reasoning are presented in Table~\ref{table_result_logical}. Our model performs the best across almost all models, including powerful ChatGPT.
This verifies that explicitly incorporating structure reasoning over text during pre-training indeed helps improve logical reasoning performance.
On the contrary, \emph{CoLAKE} and \emph{KEPLER} that incorporate external knowledge structures into language pre-training would damage the logical reasoning performance of language models in some cases.

\subsection{Ablation Study} 
\paragraph{Different Elementary Structures} To demonstrate the impact of different elementary structures on our model, we conduct an ablation study by learning structure reasoning respectively over the simple triplet, two-step path, inward-intersected pattern and outward-intersected pattern into \emph{RoBERTa}.
Table~\ref{table_ablation} presents the joint EM and joint F1 on both validation and test sets on the 2WikiMultiHopQA dataset.
The results demonstrate that the simple triplet and other three complex structures can improve multi-hop reasoning performance. 
Moreover, our model combining these four structures
performs the best, which indicates that these four types of elementary structures are complementary for complex reasoning.
\begin{table}[!th]
    \begin{center}
    \caption{Ablation study on 2WikiMultiHopQA. EM and F1 respectively denote joint EM and joint F1 scores. \emph{outward.pattern} and \emph{inward.pattern} mean the outward-intersected pattern and inward-intersected pattern.}
    \label{table_ablation}
    \resizebox{0.48\textwidth}{!}{
    \setlength{\tabcolsep}{1.7mm}{
    \begin{tabular}{l|cc|cc}
    \toprule
    \multirow{2}{*}[-0.7ex]{\centering Model}
    &\multicolumn{2}{c}{Val.} 
    &\multicolumn{2}{c}{Test} \\
    \cmidrule(lr){2-3} \cmidrule(lr){4-5} 
    & EM & F1 & EM & F1 \\
    \midrule
    \emph{RoBERTa}  & 55.53 & 62.92 & 55.20 & 62.56 \\
    \ \ \ \emph{w/ simple triplet} & 58.76 & 67.49 & 58.64 & 66.90 \\
    \ \ \ \emph{w/ two-step path} & 57.51 & 65.90 & 56.95 & 65.33  \\
    \ \ \ \emph{w/ outward.pattern} & 55.90 & 63.42 & 55.87 & 63.08  \\
    \ \ \ \emph{w/ inward.pattern} & 57.47 & 66.87 & 57.16 & 66.40  \\
    \emph{Ours} & 59.08 & 67.75 & 59.00 & 67.32   \\
    \bottomrule
    \end{tabular}
    }
    }
    \end{center}
\end{table}

\paragraph{Different Pre-training Objectives} We also provide an additional study to show how much improvement are respectively from structure reasoning loss $L_{SR}$ and masked language modeling loss $L_{MLM}$ on 2WikiMultiHopQA dataset. As shown in Table~\ref{table_ablation2}, our model simultaneously optimizing structure reasoning and masked language modeling can achieve best performance.
\begin{table}[!th]
    \begin{center}
    \caption{Ablation study of various pre-training losses on 2WikiMultiHopQA.}
    \label{table_ablation2}
    \resizebox{0.42\textwidth}{!}{
    \setlength{\tabcolsep}{1.9mm}{
    \begin{tabular}{l|cc|cc}
    \toprule
    \multirow{2}{*}[-0.7ex]{\centering Model}
    &\multicolumn{2}{c}{Val.} 
    &\multicolumn{2}{c}{Test} \\
    \cmidrule(lr){2-3} \cmidrule(lr){4-5} 
    & EM & F1 & EM & F1 \\
    \midrule
    \emph{RoBERTa}  & 55.53 & 62.92 & 55.20 & 62.56 \\
    \ \ \ \emph{w/ $L_{MLM}$} & 57.70 & 66.52 & 57.25 & 65.90 \\
    \ \ \ \emph{w/ $L_{SR}$} & 58.43 & 67.53 & 57.52 & 66.74  \\
    \emph{Ours} & 59.08 & 67.75 & 59.00 & 67.32   \\
    \bottomrule
    \end{tabular}
    }
    }
    \end{center}
\end{table}

\subsection{Further Analysis}
\label{further_analysis}
\paragraph{Transfer Ability under Low-Resource Setting} It is challenging to transfer a pre-trained model to complex reasoning tasks in low-resource settings, where only limited training data are available.
To illustrate the generalization capability of our model to the complex reasoning tasks under low-resource settings, we conduct experiments on 2WikiMultiHopQA with limited 10\% training data (see Table~\ref{table_fewshot}). 
Compared to training on the whole training set, we find that all models show a significant performance drop when only 10\% training data is available.
This suggests that fine-tuning with limited training data is nontrivial.
Furthermore, we see that our model still outperforms the others by a considerable margin, exhibiting a better low-resource transfer ability.
This is because the foundation structure reasoning skill learned by our model can help generalize to new reasoning tasks requiring structure reasoning over text.

\begin{table}[!th]
    \begin{center}
    \caption{Results on the test set of 2WikiMultiHopQA with respectively 10\% and 100\% training data.}
    \label{table_fewshot}
    \resizebox{0.48\textwidth}{!}{
    \setlength{\tabcolsep}{1.7mm}{
    \begin{tabular}{l|cc|cc}
    \toprule
    \multirow{2}{*}[-0.7ex]{\centering Model}
    &\multicolumn{2}{c}{10\%} 
    &\multicolumn{2}{c}{100\%} \\
    \cmidrule(lr){2-3}\cmidrule(lr){4-5} & Joint EM & Joint F1
    & Joint EM & Joint F1 \\
    \midrule    
    \emph{RoBERTa} & 39.28 & 49.27 & 55.20 & 62.56 \\
    \emph{CoLAKE}  & 37.69 & 47.77 & 55.25 & 62.99 \\
    \emph{KEPLER} & 39.84 & 49.49 & 54.83 & 63.00 \\
    \emph{ERICA} & 40.67 & 50.27 & 57.76 & 67.14 \\
    \emph{Ours} & 41.03 & 51.21 & 59.00 & 67.32 \\
    \bottomrule
    \end{tabular}
    }
    }
    \end{center}
\end{table}

\begin{table*}[th]
\setlength\tabcolsep{8pt}
\begin{center}
\caption{H@3 results of box embedding (BoxE) method using different pre-trained models for knowledge embedding initialization on different structured queries of FB15k dataset. `p', `i', and `u' respectively represent `relation projection', `triplet intersection', and `triplet union'.}
\label{table_result_KGreasoning}
\resizebox{0.96\textwidth}{!}{
\begin{tabular}{l|c|ccc|cc|cccc}
\toprule
Model & Avg & 1p & 2p & 3p & 2i & 3i & ip & pi & 2u & up \\
\midrule
\emph{BoxE} & 0.497 & 0.797 & 0.421 & 0.313 & 0.606 & 0.721 & 0.221 & 0.429 & 0.628 & 0.338 \\
\emph{BoxE}(RoBERTa) & 0.512 & 0.812 & 0.436 & 0.320 & 0.628 & 0.740 & 0.232 & 0.441 & 0.652 & 0.346 \\
\emph{BoxE}(CoLAKE) & 0.512 & 0.812 & 0.437 & 0.319 & 0.624 & 0.742 & 0.232 & 0.444 & 0.652 & 0.346 \\
\emph{BoxE}(KEPLER) & 0.500 & 0.799 & 0.425 & 0.318 & 0.614 & 0.727 & 0.222 & 0.424 & 0.627 & 0.341 \\
\emph{BoxE}(ERICA) & 0.511 & 0.812 & 0.435 & \bf 0.321 & 0.624 & 0.741 & 0.230 & 0.442 & 0.648 & 0.344 \\
\emph{BoxE}(Ours(PTransE)) & 0.512 & 0.812 & 0.437 & 0.318 & 0.628 & 0.742 & 0.231 & 0.444 & 0.653 & 0.343 \\
\emph{BoxE}(Ours) & \bf 0.514 & \bf 0.813 & \bf 0.439 & 0.320 & \bf 0.629 & \bf 0.745 & \bf 0.233 & \bf 0.444 & \bf 0.653 & \bf 0.346 \\
\bottomrule
\end{tabular}
}
\end{center}
\end{table*}
\paragraph{Complex KG Reasoning} 
We further conduct experiments to verify the effectiveness of our model with explicit structure reasoning skills for complex reasoning on KGs. The complex KG reasoning task aims to answer complex structural queries (first-order logic queries) on incomplete knowledge graphs with one or a set of entities. We adopt the FB15k dataset ~\cite{bordes2013translating} 
derived from Freebase for evaluation. We follow the setting of~\cite{ren2020query2box} to generate 5 types of complex structural queries (i.e., 1p, 2p, 3p, 2i, 3i) for training and 9 structural query types (i.e., 1p, 2p, 3p, 2i, 3i, ip, pi, 2u, up) for evaluation, so that we can evaluate queries that are both seen and unseen during training time. 
For a structural query, we perform explicit structure reasoning on it via step-by-step updating the geometric embedding of the query and deduce the final query box embedding.  
The entities whose embeddings are close enough to the query embedding will be predicted as the final answers.
We use different pre-trained models to initialize the embeddings of entities and relations as described in Sec~\cref{sec:structured_reasoning} and compare their reasoning performance. We fine-tune for 100,000 steps with the learning rate 1e-4. 

We report H@3 results on different structural queries of FB15k in Table~\ref{table_result_KGreasoning}. We find that box embedding method (\emph{BoxE}) with our pre-trained model achieves better performance on average. 
Our model help improves the performance on the structured queries (1p, 2p, 2i) that we have modeled during pre-training and other complex ones composed of them (3p, 3i, ip, pi).
Even on the structural queries that are unseen during the training period (ip, pi, 2u, up), our model can achieve better or comparable performance.
These observations demonstrate that our model with explicit structure reasoning is more beneficial to initialize geometric embeddings for complex KG reasoning tasks, and generalize across different knowledge structures.

\paragraph{Knowledge Probing}
To illustrate that explicit structure reasoning integrated language model can learn the structured knowledge, we conduct knowledge probing experiments on LAMA and LAMA-UHN probes. The task requires the pre-trained model to directly predict masked spans in factual descriptions without fine-tuning. The experimental results are shown in Table~\ref{table_probing}. We do not compare ERICA as it can not probe any of this factual knowledge.
We can see that although our model performs slightly worse than \emph{CoLAKE} which incorporates external surrounding knowledge in PLMs, it outperforms \emph{RoBERTa}, \emph{KEPLER} and \emph{ERICA}. This suggests that the structure reasoning over text can also help learn a certain amount of knowledge.
\begin{table}[!th]
    \begin{center}
    \caption{Knowledge probing results (P@1) on LAMA and LAMA-UHN. * indicates that the results are from~\cite{sun2020colake} and~\cite{wang2021kepler}. Others are from our implementation.}
    \label{table_probing}
    \resizebox{0.47\textwidth}{!}{
    \setlength{\tabcolsep}{1.2mm}{
    \begin{tabular}{l|cc|cc}
    \toprule
    \multirow{2}{*}[-0.7ex]{\centering Model}
    &\multicolumn{2}{c}{LAMA} 
    &\multicolumn{2}{c}{LAMA-UHN} \\
    \cmidrule(lr){2-3}\cmidrule(lr){4-5} & Google-RE & T-REx
    & Google-RE & T-REx \\
    \midrule    
    \emph{RoBERTa*} & 5.3 & 24.7 & 2.2 & 17.0 \\
    \emph{CoLAKE*}  & 9.5 & 28.8 & 4.9 & 20.4 \\
    \emph{KEPLER*} & 7.3 & 24.6 & 3.3 & 16.5 \\
    \midrule
    \emph{RoBERTa} & 4.7 & 18.9 & 1.8 & 14.5 \\
    \emph{Ours} & 8.3 & 28.8 & 3.1 & 19.2 \\
    \bottomrule
    \end{tabular}
    }
    }
    \end{center}
\end{table}

\section{Conclusion}
In this paper, we propose to inject the foundation structure reasoning skill into PLMs for complex reasoning tasks to model implicit structures within contexts and perform explicit reasoning over them.
To accomplish this objective, we present a unified framework combining structure reasoning and language modeling. It extracts four types of elementary structures from contexts to construct structured queries and adopts the stepwise embedding method for explicit structure reasoning along the constructed queries to find the answer. We utilize contextual language representations to initialize structure representations for fusing textual and structure semantics.
Experimental results show the general effectiveness of our model on complex reasoning tasks. 

In the future, we will take different complexity of knowledge structures into consideration and model a more comprehensive skill over more structures.
Besides, more foundation reasoning skills can be explored, such as abductive reasoning and external knowledge retrieval, and be combined with the structure reasoning skill for more general reasoning tasks.

\bibliographystyle{IEEEtran}
\bibliography{taslp}


 




\vfill

\end{document}